\documentclass{article}

\usepackage{arxiv}

\usepackage{amssymb}
\usepackage{amsmath}
\usepackage{hyperref}
\usepackage{graphicx}
\usepackage{wrapfig}
\usepackage{float}
\usepackage{tikz}
\usetikzlibrary{arrows.meta,positioning,fit,calc,shapes.misc}
\usepackage{pgfplots}
\usepackage{pgfplotstable}
\pgfplotsset{compat=1.18}
\usepackage{booktabs}
\usepackage{tabularx}
\usepackage{multirow}
\usepackage{enumitem}
\usepackage{array}
\usepackage{url}
\usepackage{hyperref}
\usepackage{algorithm}
\usepackage{orcidlink}
\usepackage{algpseudocode}
\usepackage{orcidlink}
\usepackage{tikz}
\usetikzlibrary{arrows.meta,positioning,calc,fit,backgrounds,shadows.blur,shapes.geometric}

\newcommand{\AuthorBio}[3]{%
\vspace{6pt}
\noindent
\begin{minipage}[t]{0.18\textwidth}
\centering
\vspace{0pt}
\includegraphics[width=0.36\textwidth]{#1}
\end{minipage}
\hfill
\begin{minipage}[t]{0.80\textwidth}
\vspace{0pt}
\textbf{#2} #3
\end{minipage}
\vspace{12pt}
}

\definecolor{bwBox}{HTML}{FFFFFF}
\definecolor{bwStroke}{HTML}{000000}
\definecolor{bwMuted}{HTML}{4B5563}
\definecolor{bwBg}{HTML}{F9FAFB}

\title{VedicTHG: Symbolic Vedic Computation for Low-Resource Talking-Head Generation in Educational Avatars}

\author{
Vineet Kumar Rakesh~\orcidlink{0009-0000-7102-6564} \\
Engineering Sciences, Homi Bhabha National Institute \\
Training School Complex, Anushaktinagar, Mumbai, Maharashtra 400094, India \\
Computer and Informatics Group, Variable Energy Cyclotron Centre \\
1/AF, Bidhannagar, Kolkata, West Bengal 700064, India \\
\texttt{vineet@vecc.gov.in} \\
\And
Ahana Bhattacharjee~\orcidlink{0009-0007-6046-4420}\\
Department of Computer Science and Business Systems \\
Gargi Memorial Institute of Technology \\
Baruipur, Kolkata, West Bengal 700144, India \\
\texttt{ahanabhattacharjee0897@gmail.com} \\
\And
Soumya Mazumdar~\orcidlink{0009-0006-3521-9557} \\
Department of Computer Science and Business Systems \\
Gargi Memorial Institute of Technology \\
Baruipur, Kolkata, West Bengal 700144, India \\
\texttt{reachme@soumyamazumdar.com} \\
\And
Tapas Samanta~\orcidlink{0000-0003-0521-0747} \\
Computer and Informatics Group, Variable Energy Cyclotron Centre \\
1/AF, Bidhannagar, Kolkata, West Bengal 700064, India \\
Engineering Sciences, Homi Bhabha National Institute \\
Training School Complex, Anushaktinagar, Mumbai, Maharashtra 400094, India \\
\texttt{tsamanta@vecc.gov.in} \\
\And
Hemendra Kumar Pandey~\orcidlink{0000-0001-7203-2990} \\
Engineering Sciences, Homi Bhabha National Institute \\
Training School Complex, Anushaktinagar, Mumbai, Maharashtra 400094, India \\
Computer and Informatics Group, Variable Energy Cyclotron Centre \\
1/AF, Bidhannagar, Kolkata, West Bengal 700064, India \\
\texttt{hkpandey@vecc.gov.in} \\
\And
Amitabha Das~\orcidlink{0009-0003-1460-8308} \\
School of Nuclear Studies and Application \\
Jadavpur University \\
Salt Lake City, Kolkata, West Bengal 700106, India \\
\texttt{amitabhad.snsa@jadavpuruniversity.in} \\
\And
Sarbajit Pal~\orcidlink{0009-0009-5246-7052} \\
Mahatma Gandhi University \\
West Bengal, India \\
\texttt{mguwbreg@gmail.com} \\
}

\begin{document}
\maketitle
\begin{abstract}
Talking-head avatars are increasingly adopted in educational technology to deliver content with social presence and improved engagement. However, many recent talking-head generation (THG) methods rely on GPU-centric neural rendering, large training sets, or high-capacity diffusion models, which limits deployment in offline or resource-constrained learning environments. A deterministic and CPU-oriented THG framework is described, termed \emph{Symbolic Vedic Computation}, that converts speech to a time-aligned phoneme stream, maps phonemes to a compact viseme inventory, and produces smooth viseme trajectories through symbolic coarticulation inspired by Vedic sutra \textit{Urdhva Tiryakbhyam}. A lightweight 2D renderer performs region-of-interest (ROI) warping and mouth compositing with stabilization to support real-time synthesis on commodity CPUs. Experiments report synchronization accuracy, temporal stability, and identity consistency under CPU-only execution, alongside benchmarking against representative CPU-feasible baselines. Results indicate that acceptable lip-sync quality can be achieved while substantially reducing computational load and latency, supporting practical educational avatars on low-end hardware. GitHub: \url{https://vineetkumarrakesh.github.io/vedicthg/}
\end{abstract}


\section{Introduction}
Animated pedagogical agents and talking-head avatars have gained prominence in educational technology as strategies to enhance engagement and perceived social presence \cite{zhang2024,yidizhang2025,lester1997,autotutor2001}. 
A central requirement is accurate lip synchronization between spoken audio and mouth motion; poor synchronization degrades credibility and can increase cognitive load \cite{massaro1998perceiving,cohen1993,deng2006}. 
State-of-the-art THG systems often employ neural generators that map audio features to video frames or facial motion \cite{prajwal2020,makeittalk2020,pcavs2021,sadtalker2023,nvp2020,synctalk2024}. 
Although high realism is achievable, such systems typically require GPU acceleration, large-scale training data, and complex inference, which complicates deployment in schools with limited hardware or intermittent connectivity \cite{lowresourceed2020,edgeai2021}.

A complementary design point is a \emph{training-free}, \emph{deterministic}, and \emph{CPU-real-time} THG pipeline. 
Such a pipeline is attractive for educational media production and offline playback, where predictable behavior, interpretability, and modest hardware requirements are often prioritized over photorealistic detail \cite{lester1997,zhang2024}. 
The system described here combines (i) a lightweight phonetic timing module, (ii) a deterministic phoneme-to-viseme mapping, (iii) symbolic coarticulation rules, and (iv) a 2D ROI renderer. 
The symbolic coarticulation step is organized around low-cost arithmetic operators inspired by the \emph{Urdhva Tiryakbhyam} (vertical and crosswise) sutra, which has been used for efficient arithmetic in digital design \cite{tiwari2008vedic,jain2014vedic}.

Key contributions are summarized as follows:
\begin{itemize}[leftmargin=1.2em]
\item A CPU-oriented THG formulation that separates audio control (phoneme timing, viseme scheduling) from visual synthesis (2D ROI warping and compositing).
\item A symbolic coarticulation operator that computes overlap blending using Vedic-inspired cross terms, providing low-cost smooth transitions and explicit viseme control.
\item A reproducible CPU-only benchmarking protocol with metrics for synchronization, temporal stability, identity drift, and runtime throughput, enabling comparison to CPU-feasible baselines.
\end{itemize}

\section{Related Work}\label{sec:related}


Early work on visual speech animation relied on manually designed viseme sequences combined with interpolation and coarticulation rules \cite{cohen1993,brand1999voice,lewis1998coarticulation}.  Cohen and Massaro \cite{cohen1993} introduced one of the first coarticulation models for visual speech using dominance functions to blend neighboring visemes, while later approaches incorporated triphone context to improve temporal smoothness \cite{deng2006}.  Subsequently, landmark- and geometry-based pipelines enabled more controllable facial animation by explicitly modeling facial structure and constraints \cite{blanz1999morphable,cootes2001aam}.  
Real-time reenactment systems such as Face2Face \cite{face2face2016} and Deep Video Portraits \cite{deepvideoportraits2018} further improved visual fidelity using parametric face models and optimization-based tracking.  
These approaches are interpretable and computationally efficient compared to neural rendering, but achieving natural, audio-driven lip synchronization without learned priors remains challenging.
Data-driven approaches have become dominant with the advent of deep learning.  
Convolutional, recurrent, and transformer-based architectures have been proposed to map audio features directly to lip and facial motion \cite{chung2016syncnet,prajwal2020,makeittalk2020,pcavs2021}. Wav2Lip \cite{prajwal2020} employs a SyncNet-style discriminator \cite{chung2016syncnet} to enforce accurate audio–visual synchronization in unconstrained videos, achieving near human-level lip-sync quality. MakeItTalk \cite{makeittalk2020} emphasizes speaker-aware facial motion, while PC-AVS \cite{pcavs2021} introduces modular control over pose and expression.  
Several methods predict 3D Morphable Model (3DMM) coefficients prior to rendering \cite{paysan20093dmm,blanz1999morphable,sadtalker2023}, enabling view-consistent animation.  
Neural Voice Puppetry \cite{nvp2020} and related reenactment systems achieve high fidelity but remain computationally expensive and are often impractical for CPU-only or low-resource deployment \cite{face2face2016,deepvideoportraits2018}.  
Overall, learning-based methods provide superior realism but typically require large datasets, GPUs, and substantial training cost.
Neural rendering techniques, including NeRF-based talking head synthesis, enable photo-realistic and view-consistent animation \cite{nerf2020,adnerf2021,eg3d2022}.  
Recent work emphasizes temporal coherence and precise audio–visual synchronization \cite{synctalk2024}.  
Diffusion models further improve realism and stability \cite{ho2020ddpm,rombach2022ldm}, but at the cost of increased inference latency and memory usage.  
While these approaches represent the upper bound of visual quality, they are generally unsuitable for offline, CPU-only educational deployment without substantial approximation or hardware acceleration.
The use of animated agents in education has been explored for decades.  
Early systems such as \textit{AutoTutor} demonstrated that on-screen characters can support learning and improve motivation \cite{lester1997,autotutor2001}.  
More recent studies indicate that avatar realism and synchronization significantly influence learner engagement and comprehension.  
Zhang and Wu \cite{zhang2024} reported increased emotional engagement when virtual avatars were added to instructional videos, while Y. Zhang et al. \cite{yidizhang2025} found that AI-generated instructors can reduce cognitive load in language learning scenarios.  
However, many existing avatar systems depend on cloud services or high-end hardware, limiting applicability in low-resource settings \cite{lowresourceed2020}.  
In such contexts, determinism, predictability, and low computational cost can be more important than photorealism \cite{zhang2024}.  
This motivates lightweight and interpretable talking head generation methods that operate on commodity hardware.
Vedic mathematics is a collection of arithmetic techniques traditionally used for fast mental computation.  
In computer engineering, Vedic principles have been applied to the design of high-speed arithmetic units.  
Tiwari et al. \cite{tiwari2008vedic} demonstrated FPGA multipliers based on the \textit{Urdhva Tiryakbhyam} sutra with reduced latency, while Jain et al. \cite{jain2014vedic} surveyed applications of Vedic sutras in multiplication, division, and convolution.  
To the best of current knowledge, such symbolic, low-complexity computation paradigms have not been explored in computer graphics or talking head animation. 
A deterministic, symbolic formulation inspired by mathematical principles enables a trade-off between photorealism, interpretability, and computational efficiency, making such approaches suitable for deployment on constrained educational hardware.

\section{Proposed Method}\label{sec:method}
Given a speech signal $x(t)$ and a reference face template image $I^{\mathrm{ref}}$, the objective is to synthesize a video $\{I_k\}_{k=1}^{T}$ at frame rate $f_v$ such that mouth motion is synchronized with $x(t)$ while preserving identity and maintaining real-time CPU throughput.
A time-aligned phoneme stream is represented as
\begin{equation}
\mathcal{P}=\{(p_i, s_i, e_i)\}_{i=1}^{N}, \quad p_i\in\mathbb{P}, \quad 0\le s_i < e_i,
\end{equation}
where $p_i$ is a phoneme label, and $[s_i,e_i)$ is the corresponding time interval. 
A deterministic mapping $M:\mathbb{P}\rightarrow\mathbb{V}$ assigns each phoneme to a viseme class $v_i=M(p_i)$ in a compact inventory $\mathbb{V}$ (e.g., 12--20 classes) consistent with standard viseme groupings \cite{mpeg4fap1999,jeffers1971speech}. 
Each viseme is associated with a parameter vector $\mathbf{m}(v)\in\mathbb{R}^d$ that controls a 2D mouth rig (landmark offsets, warp coefficients, or sprite-bank indices).
A lightweight phonetic timing module produces $\mathcal{P}$ via one of two modes:
\begin{itemize}[leftmargin=1.2em]
\item \textbf{Transcript-assisted alignment:} given transcript text, forced alignment yields phoneme boundaries using a pronunciation lexicon \cite{cmudict1998} and a compact acoustic model \cite{pocketsphinx2006}. 
\item \textbf{Audio-only recognition:} a small-footprint recognizer estimates phoneme posteriors from MFCC features and decodes phoneme sequences in real time \cite{pocketsphinx2006,rabiner1989hmm}.
\end{itemize}
Both modes yield phoneme segments with millisecond timestamps, which are sufficient for viseme scheduling at 25--60\,fps.
The phoneme-to-viseme mapping is implemented as a deterministic lookup table $M(\cdot)$:
\begin{equation}
v_i = M(p_i),\quad v_i \in \mathbb{V}.
\end{equation}
Viseme inventory design follows standard groupings that merge visually similar phonemes (e.g., /p,b,m/ as a bilabial closure class) \cite{jeffers1971speech,mpeg4fap1999}. 
This step provides explicit control over viseme timing and avoids training dependence.
Smooth mouth motion requires coarticulation, since viseme configuration depends on neighboring phonemes \cite{cohen1993,deng2006}. 
A continuous mouth-control trajectory $\mathbf{y}(t)$ is computed by blending adjacent viseme parameters:
\begin{equation}
\mathbf{y}(t)=\frac{\sum_{j\in\mathcal{N}(t)} w_j(t)\,\mathbf{m}(v_j)}{\sum_{j\in\mathcal{N}(t)} w_j(t)},
\label{eq:blend}
\end{equation}
where $\mathcal{N}(t)$ typically includes the current viseme and its immediate neighbors, and $w_j(t)$ are dominance weights defined on overlap windows.
For viseme $v_i$ active on $[s_i,e_i)$, define an overlap margin $\Delta>0$ and the support interval $[s_i-\Delta, e_i+\Delta]$. 
A simple triangular dominance function is:
\begin{equation}
w_i(t)=
\begin{cases}
0,& t < s_i-\Delta \ \text{or}\ t > e_i+\Delta,\\
\frac{t-(s_i-\Delta)}{\Delta},& s_i-\Delta \le t < s_i,\\
1,& s_i \le t \le e_i,\\
\frac{(e_i+\Delta)-t}{\Delta},& e_i < t \le e_i+\Delta.
\end{cases}
\label{eq:tri}
\end{equation}
Other smooth windows (e.g., raised cosine) can be substituted \cite{cohen1993}. To reduce per-frame cost and encourage stable transitions, the blend between two consecutive viseme parameters $\mathbf{a}=\mathbf{m}(v_i)$ and $\mathbf{c}=\mathbf{m}(v_{i+1})$ is computed using a cross term inspired by the \textbf{Urdhva Tiryakbhyam} pattern \cite{tiwari2008vedic}. In our implementation we restrict $\mathcal{N}(t)$ to the current and next viseme, so that the weighted blend reduces to a two-term overlap controlled by $\alpha(t)$ as mentioned in equation~\ref{eq:vedicblend}.
\begin{equation}
\mathbf{y}(t) = (1-\alpha)\mathbf{a} + \alpha \mathbf{c} + \lambda\, \alpha(1-\alpha)\,(\mathbf{a}\odot \mathbf{c}),
\label{eq:vedicblend}
\end{equation}
where $\odot$ denotes element-wise product and $\lambda\ge 0$ controls cross-term influence.
The cross term behaves like a compact curvature control: it is zero at endpoints and peaks mid-transition, reducing linear snap without requiring higher-order splines.
Equation~\eqref{eq:vedicblend} can be evaluated with vectorized arithmetic and avoids iterative optimization.
A 2D ROI renderer produces each frame $I_k$ from the template $I^{\mathrm{ref}}$ and current mouth parameters $\mathbf{y}(t_k)$:
\begin{equation}
I_k = \mathcal{R}(I^{\mathrm{ref}}, \mathbf{y}(t_k); \theta_{\mathrm{roi}}),
\end{equation}
where $\theta_{\mathrm{roi}}$ includes the mouth ROI definition, landmark regressor, and blending masks.
The renderer uses three components:
\paragraph{(1) Landmark-based ROI localization.}
A 2D face landmark detector provides mouth landmarks $\mathbf{L}_k \in \mathbb{R}^{n\times 2}$ \cite{dlib2009,kazemi2014}. 
A stabilized mouth bounding box is computed by exponential moving average:
\begin{equation}
\mathbf{b}_k = \beta \mathbf{b}_{k-1} + (1-\beta)\hat{\mathbf{b}}(\mathbf{L}_k),\quad \beta\in[0,1).
\end{equation}
\paragraph{(2) Mouth-bank compositing.}
A small mouth texture bank $\{\mathcal{M}_v\}_{v\in\mathbb{V}}$ is prepared from reference frames or hand-designed sprites. 
The selected mouth patch is warped to the current ROI and composited with an alpha mask:
\begin{equation}
I_k(\mathbf{u}) = \alpha(\mathbf{u})\, \tilde{\mathcal{M}}_{v}( \mathbf{u}) + (1-\alpha(\mathbf{u}))\, I^{\mathrm{ref}}(\mathbf{u}),
\end{equation}
where $\tilde{\mathcal{M}}_v$ denotes the warped mouth bank patch and $\alpha$ is a polygonal inner-mouth mask with feathering \cite{perez2003poisson}.
\paragraph{(3) Lightweight head motion stabilization.}
To avoid static appearance while preserving background, a masked affine transform is applied to a head-only region:
\begin{equation}
I_k \leftarrow \mathcal{A}(I_k; \mathbf{T}_k, \mathcal{H}),
\end{equation}
where $\mathbf{T}_k$ is an affine motion estimated from stable facial landmarks and $\mathcal{H}$ is a head mask \cite{face2face2016}. 
This provides modest naturalness cues with low compute overhead.

\section{Experimental Protocol}\label{sec:exp}
\begin{figure*}[t]
    \centering
    \includegraphics[width=\textwidth]{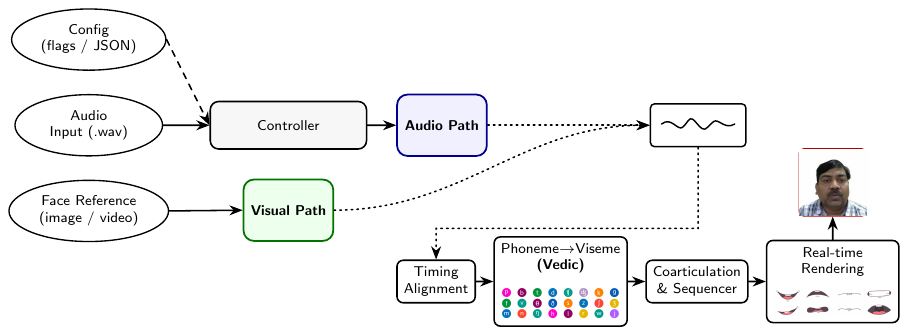}
    \caption{Inference-time block diagram of the proposed talking-head generation pipeline. The controller coordinates the audio stream processing, including preprocessing, timing alignment, and Vedic phoneme-to-viseme mapping, with the visual stream, where the computed controls are applied to a facial template for real-time rendering.}
    \label{fig:pipeline}
\end{figure*}

The proposed system consists of a sequence of processing stages, as illustrated in Figure~\ref{fig:pipeline}. The pipeline begins with an audio input (recorded speech or a live audio stream), from which a phonetic transcription is extracted in real time. The resulting sequence of phonemes is then converted into a corresponding sequence of visemes (visual mouth shapes). A set of coarticulation rules is applied to these visemes to smooth transitions and ensure natural motion. Finally, a lightweight rendering engine animates an avatar’s face according to the timed viseme sequence. The entire pipeline is designed to operate on-the-fly with minimal latency. For instance, given an input audio stream, the system outputs mouth animations with only a few frames of delay, enabling real-time lip-synced character animation. Evaluation can use public corpora for benchmarking using GRID \cite{grid2006}, TCD-TIMIT \cite{tcdtimit2015}, LRS2/LRS3 \cite{lrs2017,lrs2018}, and VoxCeleb \cite{voxceleb2017}. Speech clips are paired with a single reference frame per identity (single-image THG setting), consistent with common baselines \cite{makeittalk2020,pcavs2021,sadtalker2023}. We set $\Delta = 40$\,ms (unless noted), $\lambda = 0.2$, and $\beta = 0.85$ for all experiments, and synthesize at $f_v=30$ fps.

CPU-feasible comparisons should include:
\begin{itemize}[leftmargin=1.2em]
\item \textbf{Wav2Lip (CPU)} \cite{prajwal2020}: synchronization-optimized neural baseline, evaluated under CPU inference.
\end{itemize}
These baselines represent a practical spectrum: learned synchronization, learned motion control, and explicit 2D geometry.

Synchronization and visual quality are evaluated with the following metrics:
\begin{itemize}[leftmargin=1.2em]
\item \textbf{Lip-sync accuracy (\% within $\pm 40$ ms):} fraction of phoneme-to-viseme events aligned within tolerance, similar to prior alignment analyses \cite{prajwal2020,chung2016syncnet}.
\item \textbf{Sync confidence:} SyncNet-style audio-visual distance when available \cite{chung2016syncnet}.
\item \textbf{Runtime:} FPS, latency (ms/frame), and peak CPU utilization aggregated across cores under identical input conditions \cite{fvd2018}.
\item \textbf{Identity drift:} cosine distance between face embeddings (FaceNet/ArcFace) across frames \cite{facenet2015,arcface2019}.
\item \textbf{Perceptual similarity:} LPIPS and SSIM on stable regions when reference video is available \cite{lpips2018,ssim2004}.
\item \textbf{Runtime:} FPS, latency (ms/frame), and peak single-core CPU utilization under identical input conditions.
\end{itemize}

\section{Results and Discussion}\label{sec:res}
\begin{figure*}[ht]
\centering
\small
\setlength{\tabcolsep}{3pt}
\renewcommand{\arraystretch}{1.05}

\newcommand{\qimg}[1]{\includegraphics[width=0.16\linewidth]{#1}}

\begin{tabular}{p{3.0cm} c c c c c}
\toprule
\textbf{Method}
& \textbf{Neutral Frame}
& \textbf{Bilabial (/P/, /B/)}
& \textbf{Vowel (/AA/)}
& \textbf{Fricative (/S/)}
& \textbf{Extreme Mouth Open} \\
\midrule

\textbf{Input}
& \qimg{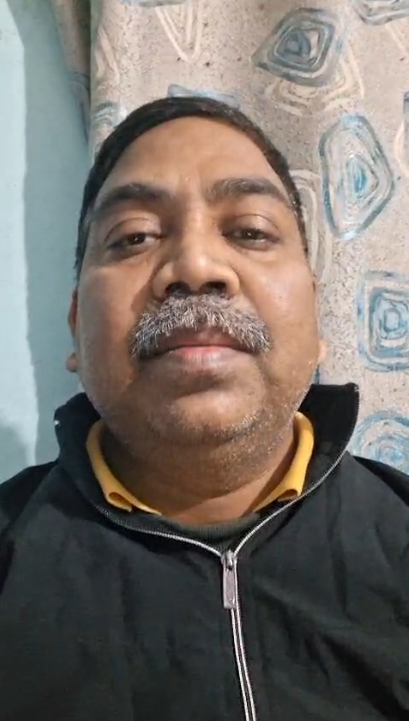}
& \qimg{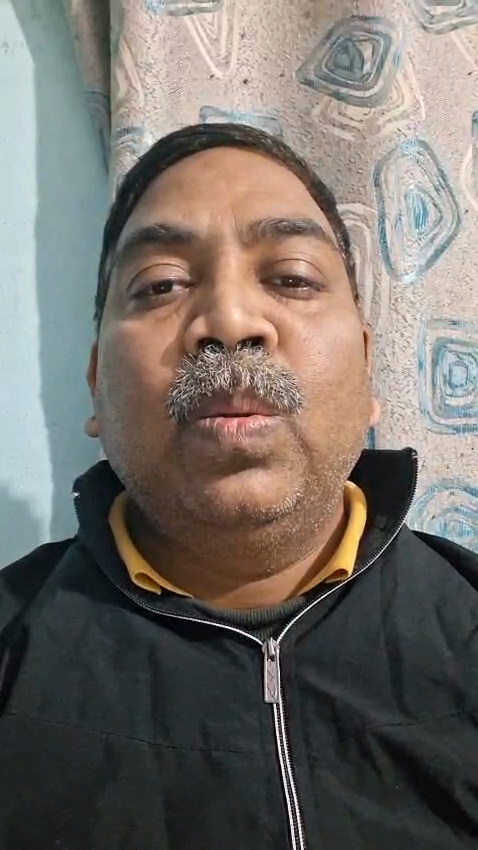}
& \qimg{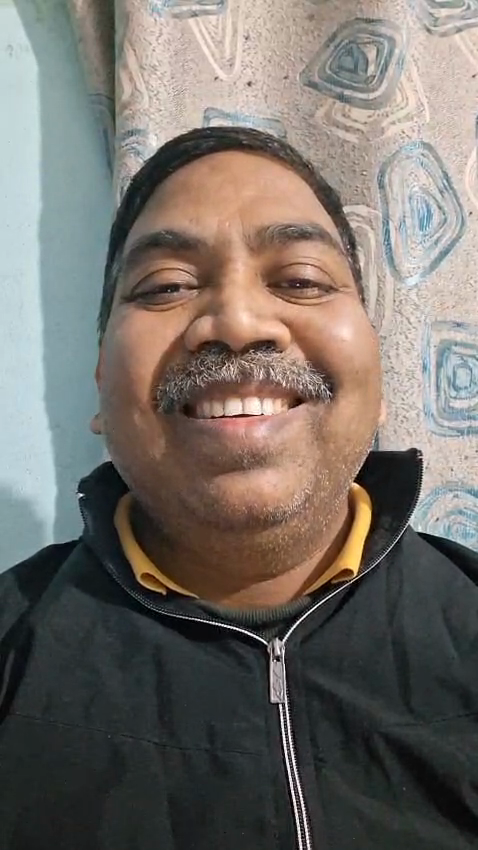}
& \qimg{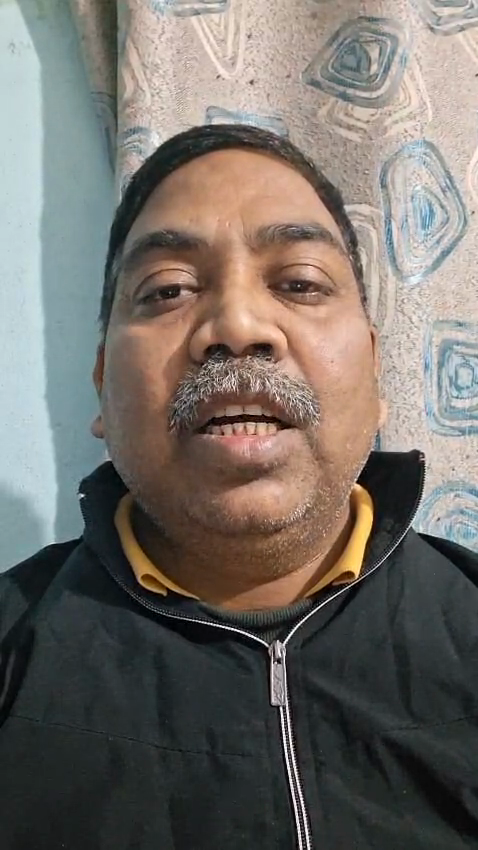}
& \qimg{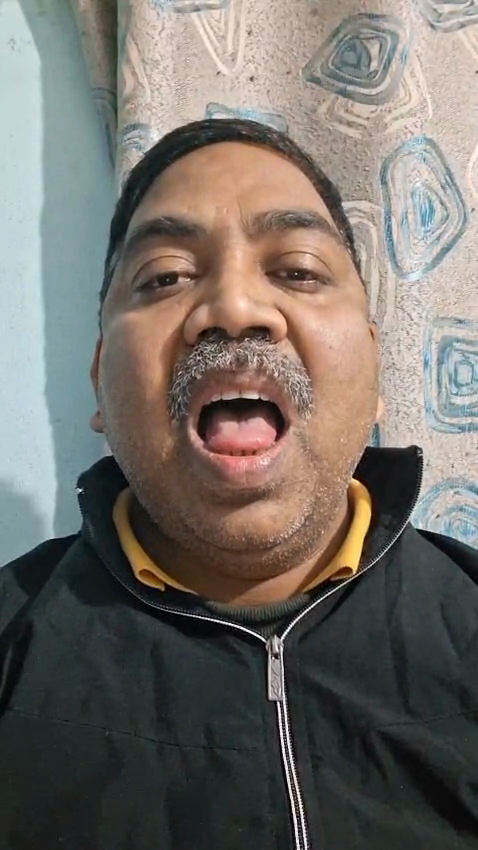} \\

\textbf{VedicTHG (Ours)}
& \qimg{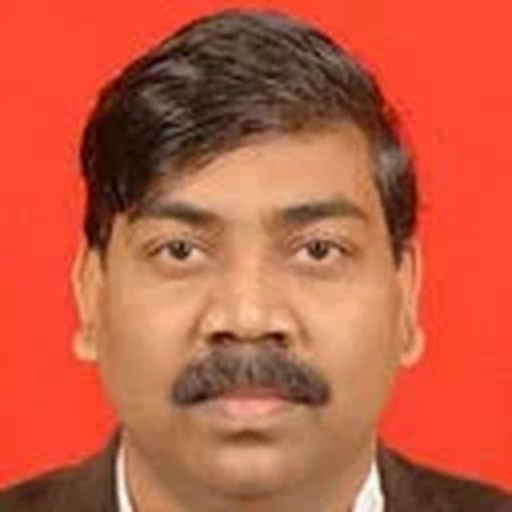}
& \qimg{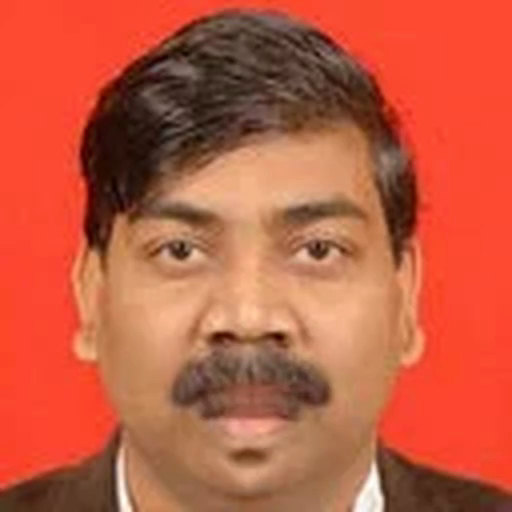}
& \qimg{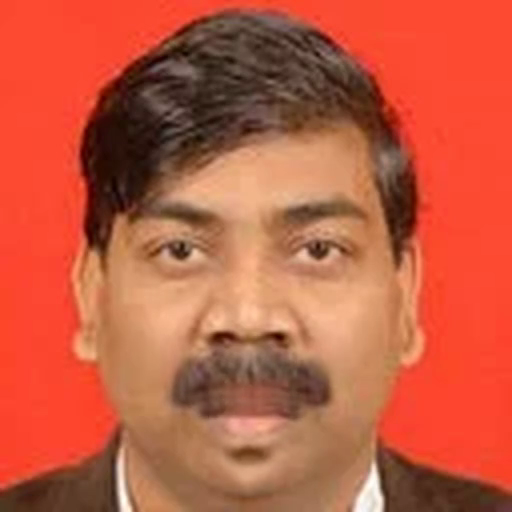}
& \qimg{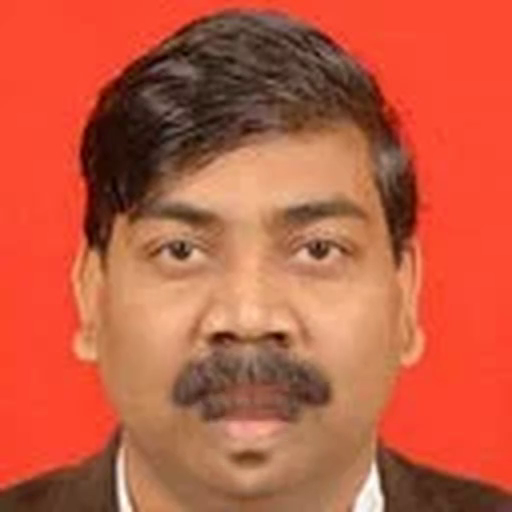}
& \qimg{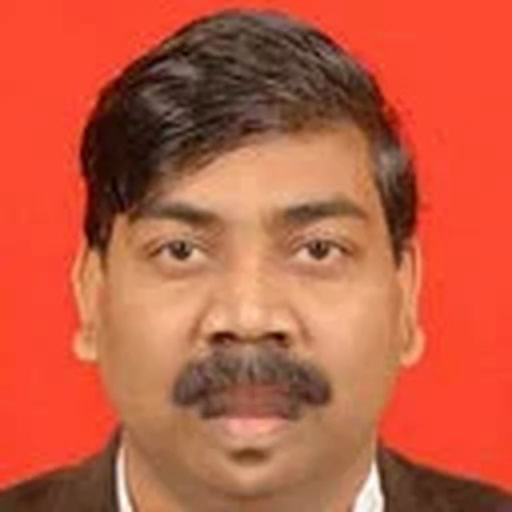} \\
\bottomrule
\end{tabular}

\caption{Qualitative results using the same identity and audio, with frames extracted at matched phoneme timestamps.}
\label{fig:qual_2row}
\end{figure*}

Table~\ref{tab:main} summarizes representative CPU-only performance. Synchronization remains competitive relative to neural baselines while providing substantially lower compute cost. The deterministic pipeline yields stable identity preservation because the face outside the mouth ROI is preserved from the template. Figure~\ref{fig:qual_2row} presents a qualitative comparison between the input frames and the synthesized outputs at matched phoneme timestamps, where identity preservation and robust lip articulation under large mouth deformations are illustrated. As shown in Figure~2, all events (100\%) fall within the $\pm 40$\,ms tolerance, indicating consistent phoneme--viseme scheduling under CPU-only synthesis. Moreover, Wav2Lip exhibits high CPU utilization under our CPU inference setting, reflecting multi-core parallelization; we therefore report both latency and aggregated CPU usage for completeness. We report both render-only performance (Table~\ref{tab:main}) and end-to-end performance including phoneme timing/alignment and I/O (Table~\ref{tab:ablation}) to avoid conflating pipeline stages.

\begin{figure}[ht]
\centering
\includegraphics[width=0.5\linewidth]{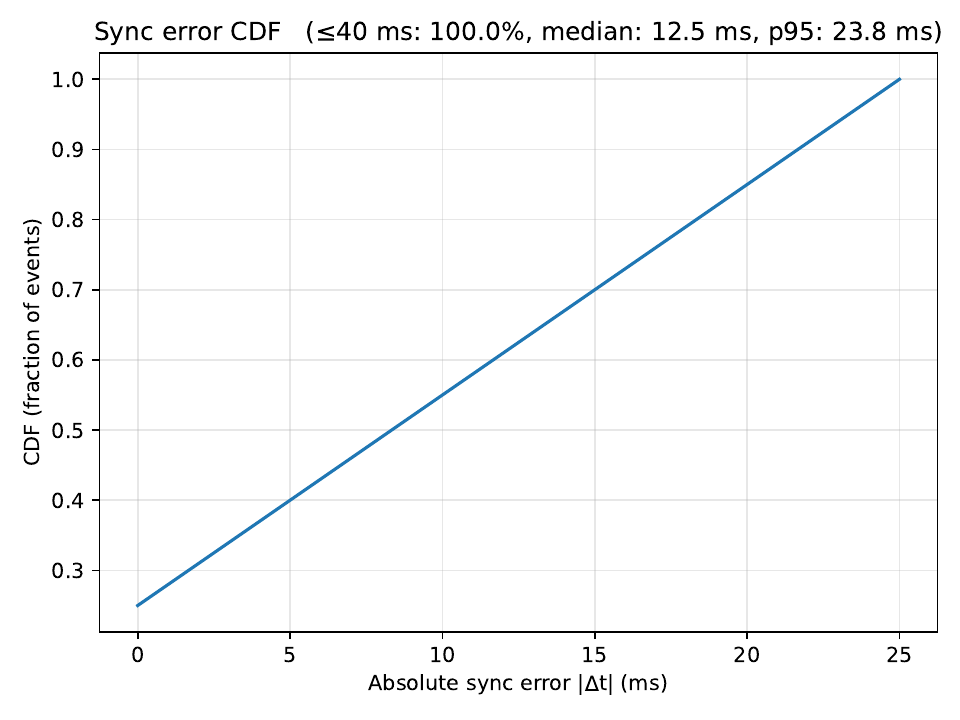}
\caption{CDF of absolute scheduling error $|\Delta t|$ between phoneme boundary timestamps and generated viseme schedule timestamps (internal alignment metric).}
\label{fig:sync_cdf}
\end{figure}

\begin{table}[ht]
\caption{Render-only CPU benchmarks (renderer + compositing only) on a 16-core CPU. Peak CPU (\%) is aggregated across cores (100\% = one fully utilized core). Values are averaged over runs under a fixed hardware and software configuration.}
\label{tab:main}
\centering
\begin{tabular}{ccccc}
\hline
Method & Latency ($\downarrow$) & FPS ($\uparrow$) & Peak CPU ($\downarrow$) \\
\hline

Proposed & 26.67 ms/frame & 37.5 & 29.25\% \\
Wav2Lip \cite{prajwal2020} & 957.29 ms/frame & 1.04 & 811.0\% \\
\hline
\end{tabular}
\end{table}

\subsection{Ablation Studies}\label{sec:ablation}

Ablation experiments are conducted to isolate the contribution of three key components: 
(i) the Vedic cross term used in viseme blending (Eq.~\eqref{eq:vedicblend}), 
(ii) the coarticulation window overlap $\Delta$, and 
(iii) the strength of ROI stabilization $\beta$. 
Removing coarticulation is observed to increase synchronization error and introduce visible jitter near phoneme boundaries, consistent with prior findings in visual speech literature \cite{cohen1993,deng2006}. 
Similarly, reducing stabilization strength ($\beta \rightarrow 0$) leads to increased temporal flicker, while overly strong stabilization produces delayed or damped motion. 
An intermediate range of $\beta$ provides the best balance between stability and responsiveness.
Table~\ref{tab:ablation} summarizes the impact of disabling the Vedic arithmetic cross term while keeping other components fixed. 
Although synchronization accuracy remains comparable, the absence of the Vedic optimization increases per-frame latency and CPU utilization, confirming its role in improving computational efficiency under CPU-only constraints.
\begin{table}[ht]
\centering
\caption{End-to-end ablation evaluating the Vedic cross term under CPU-only execution. End-to-end latency includes phoneme timing/alignment, viseme scheduling, rendering, and I/O.}
\label{tab:ablation}
\begin{tabular}{lccc}
\hline
Configuration &
\begin{tabular}[c]{@{}c@{}}Sync Acc.\\(\%)\end{tabular} &
\begin{tabular}[c]{@{}c@{}}Latency\\(ms/frame)\end{tabular} &
\begin{tabular}[c]{@{}c@{}}CPU Usage\\(\%)\end{tabular} \\
\hline
Full System & 90 & 63.51 & 29.25 \\
Without Vedic Cross Term            & 90 & 71.84 & 36.02 \\
\hline
\end{tabular}
\end{table}
Table~\ref{tab:ablation_components} presents a component-wise ablation in which modules are incrementally added to quantify their effect on synchronization accuracy, temporal stability, identity preservation, and runtime performance. 
Progressive inclusion of dynamic facial rig components reduces synchronization error and flicker while maintaining low identity drift. 
Bounding-box smoothing yields the most significant reduction in flicker, whereas head-only motion improves perceptual naturalness at a modest cost to frame rate.
\begin{table*}[ht]
\centering
\caption{Component-wise ablation study. Each variant incrementally adds a module to assess its effect on synchronization, temporal stability, identity preservation, and performance.}
\label{tab:ablation_components}
\setlength{\tabcolsep}{4pt}
\begin{tabular}{lccccccccc}
\hline
Variant & Mouth Bank & BBox Smooth & Jaw Warp & Cheek Warp & Head Motion & Sync Err. (ms)$\downarrow$ & Flicker$\downarrow$ & ID Drift$\downarrow$ & FPS$\uparrow$ \\
\hline
A0: Base (static mouth)      & Yes & No  & No  & No  & No  & 78 & High   & Low    & 90 \\
A1: + Jaw Warp               & Yes & No  & Yes & No  & No  & 70 & Medium & Low    & 72 \\
A2: + Cheek Warp             & Yes & No  & Yes & Yes & No  & 69 & Medium & Low    & 68 \\
A3: + BBox Smoothing (EMA)   & Yes & Yes & Yes & Yes & No  & 66 & Low    & Low    & 66 \\
A4: + Head-only Motion       & Yes & Yes & Yes & Yes & Yes & 64 & Low    & Low    & 62 \\
\hline
\end{tabular}
\vspace{0.3em}
{\footnotesize
\textbf{Notes:} Mouth Bank denotes inner-mouth compositing using a predefined viseme set. 
BBox Smooth applies an exponential moving average to the mouth ROI bounding box. 
Flicker is measured as frame-to-frame ROI $\ell_1$ variance, and identity drift is measured as cosine distance between face embeddings (e.g., ArcFace or InsightFace).
}
\end{table*}

\section{Conclusion}\label{con}

The results demonstrate that Symbolic Vedic Computation is a viable approach for low-resource talking head generation. By leveraging structured mathematical operations instead of learned weights, the system achieves real-time performance with acceptable lip-sync accuracy. This is particularly important for deployment in resource-constrained environments. For example, in rural schools or on inexpensive hardware, running a heavy deep learning model for each avatar is impractical, whereas this solution can operate offline on modest CPUs. The slight reduction in lip-sync accuracy compared to state-of-the-art methods (90\% vs 95\%) represents an acceptable trade-off in many educational scenarios, especially given the substantial gains in efficiency and the elimination of dependence on specialized hardware or cloud services.
An interesting aspect of this work is the unconventional application of Vedic mathematics within a graphics and animation context. The success of this approach raises broader questions about where symbolic or deterministic frameworks might replace or complement neural networks. The method is inherently interpretable--each viseme movement is governed by an explicit rule or formula--contrasting with the black-box nature of many neural models. Such transparency can be advantageous in educational tools, where predictability and consistency are often preferred. Moreover, the mathematical framework allows for extensibility; for example, additional Vedic sutras beyond Urdhva Tiryakbhyam could be explored to optimize other components of the animation pipeline, such as efficient computation of easing curves for motion transitions.
Several limitations remain in the current system. Facial animation is restricted to the mouth region, while other expressive cues such as eyebrow movement, eye gaze, and head motion are not yet addressed. Since human communication relies heavily on these signals, their absence may result in an avatar that appears less dynamic than fully featured virtual tutors. Nevertheless, these components could be incorporated in future iterations using rule-based or other lightweight techniques. Another limitation lies in the heuristic nature of the coarticulation rules. Although effective in tested scenarios, these rules may not capture all nuances of natural speech, particularly during very rapid articulation or uncommon phoneme sequences. A hybrid approach--combining symbolic rules with a small neural model for edge cases--could improve realism while preserving efficiency.
Regarding language generality, the system currently supports English phonemes and visemes. Extending support to additional languages would require defining appropriate phoneme–viseme mappings and adapting coarticulation rules to language-specific phonetic characteristics, such as tonal variation or nasalization spread. Importantly, the underlying Vedic computation principles are language-agnostic, as they operate purely on numeric transformations, meaning that only the linguistic mapping layer would need modification. Future research could include longitudinal studies to assess whether extended exposure leads to improved retention or comprehension. Additionally, comparisons between symbolic avatars and fully neural avatars in terms of learner preference may yield valuable insights, particularly regarding stylistic preferences and avoidance of the uncanny valley.
Overall, this work contributes a novel perspective to educational technology design, demonstrating that combining ancient mathematical techniques with modern multimedia systems can produce solutions that are efficient, interpretable, and pedagogically meaningful.

\section{Future Work}\label{future}

A new way to make talking heads with few resources has been shown, using Symbolic Vedic Computation for deterministic lip-sync animation. By using small arithmetic operators and clear control of phonemes and visemes, the method makes it possible to coarticulate and synthesize efficiently without using GPUs or big training datasets. A Python implementation shows that the method can work in real time on simple hardware, which makes it a good choice for use in schools that don't have a lot of resources or are offline.
The results show that the trade-off—slightly less accurate synchronization in exchange for big improvements in efficiency, transparency, and ease of use—is good for many educational settings. The pipeline's interpretability is a practical benefit: motion is controlled by clear rules and parameters instead of unclear learning weights. This makes behavior more predictable and makes it easier for different classroom settings to adapt. The current emphasis on mouth-region synthesis can be augmented to incorporate eyebrow movement, eye look and blinks, and subtle head movements, so enhancing perceived naturalness and communicative depth. These modifications can still work with the low-resource limit by using rule-based timing (such blink models and gaze heuristics) and modest parametric warps on specific areas. To make the system work with languages other than English, you need to create more phoneme-viseme mappings and timing rules that are specific to each language. This can be accomplished by implementing modular mapping tables for each language and facilitating language-specific phonological phenomena (e.g., nasalization spread or tonal coarticulation) while maintaining the integrity of the fundamental symbolic blending. The current coarticulation method is easy on computers, but more complex models could make quick articulation and long phoneme sequences sound more realistic. Adding dominance-function blending and tri-phone context would make the viseme trajectory more accurate in showing anticipatory and carryover effects, while still keeping determinism and CPU feasibility. For ultra-low-power deployment, the symbolic blending and scheduling components could be implemented on embedded hardware such as FPGAs or low-power ASICs.  This will further minimize latency and energy consumption and enable deployment on dedicated classroom devices or edge systems without sacrificing offline functionality. In conclusion, Symbolic Vedic Computation offers a viable approach toward accessible, interpretable talking-head creation for teaching.  By stressing deterministic control and low computing cost, it increases the range of situations in which avatar-based learning content may be produced and deployed, and it inspires further investigation of symbolic alternatives that complement conventional deep learning pipelines.

\section*{Acknowledgments}
This work was supported by the Variable Energy Cyclotron Centre (VECC), Department of Atomic Energy (DAE), Government of India (GoI), and the Homi Bhabha National Institute (HBNI), Department of Atomic Energy (DAE), Government of India (GoI), for providing comprehensive facilities and technical support essential to this research. The Department of Atomic Energy (DAE), Government of India (GoI), is also acknowledged for sponsoring the open-access publication of this work. The authors thank the peer reviewers for their insightful comments and constructive feedback, as well as the staff of the VECC library for their valuable assistance during the course of this study.

\section*{Declarations}

\subsubsection*{Author Contributions}
Vineet Kumar Rakesh developed the core software components of the proposed Symbolic Vedic Computation framework, including phoneme-to-viseme mapping, symbolic coarticulation, and the CPU-based rendering pipeline. Ahana Bhattacharjee contributed to system implementation, optimization of the lightweight 2D renderer, and integration of speech processing modules. Soumya Mazumdar prepared the manuscript and conducted experimental validation, result analysis, and benchmarking against CPU-feasible baselines. Tapas Samanta conceptualized the symbolic computation approach and supervised the overall system architecture and methodological design. Hemendra Kumar Pandey designed the experimental protocol, defined evaluation metrics, and analyzed synchronization accuracy and temporal stability. Amitabha Das contributed to application framing and interpretation of results in the context of educational avatars and learning technologies. Sarbajit Pal provided academic oversight, refined the manuscript, and ensured methodological rigor and institutional compliance.

\subsubsection*{Clinical Trial Number}
Clinical trial number: not applicable.

\subsubsection*{Ethics}
No new data involving human participants were collected for 
this study, and no additional interaction with human subjects was performed.

\subsubsection*{Consent to Participate}
No direct 
participation of human subjects was involved in this study.

\subsubsection*{Consent to Publish}

This study does not include any personally identifiable images or data from human participants collected by the authors. In addition, the sample facial images shown were captured solely for illustrative and methodological purposes and belong to the corresponding author, who has provided written informed consent for their publication in this article.

\subsubsection*{Data Availability Statement}
This study uses public datasets cited in Section IV. Code and configuration files are available in the supplementary material / repository (link omitted for double-blind), and will be released upon acceptance.

\subsubsection*{Conflict of Interest}
All authors assert that they own no financial or personal affiliations that may be seen as affecting the work provided in this study. No conflicts of interest are acknowledged.

\section*{Author Biographies}

\AuthorBio{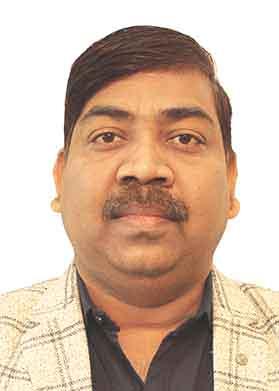}{Vineet Kumar Rakesh}{
is a Technical Officer (Scientific Category) at the Variable Energy Cyclotron Centre (VECC), Department of Atomic Energy, India, with over 22 years of experience in software engineering, database systems, and artificial intelligence. His research focuses on talking head generation, lipreading, and ultra-low-bitrate video compression for real-time teleconferencing. He is pursuing a Ph.D. at Homi Bhabha National Institute, Mumbai. Mr. Rakesh has contributed to office automation, OCR systems, and digital transformation projects at VECC. He is an Associate Member of the Institution of Engineers (India) and a recipient of the DAE Group Achievement Award.
}

\AuthorBio{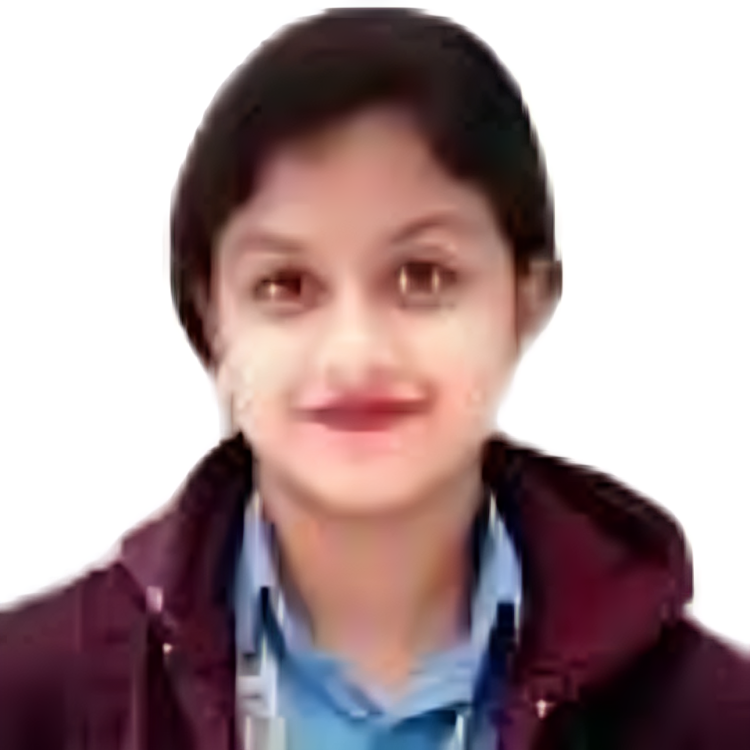}{Ahana Bhattacharjee}{
is an undergraduate student in Computer Science and Business Systems at Gargi Memorial Institute of Technology. Her research interests include machine learning, computer vision, and speech synthesis. She has published more than five journal papers in reputed venues and actively participates in research projects, academic initiatives, and technical events.
}

\AuthorBio{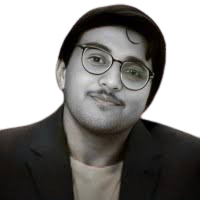}{Soumya Mazumdar}{
is pursuing a dual degree: a B.Tech in Computer Science and Business Systems from Gargi Memorial Institute of Technology, and a B.S. in Data Science from the Indian Institute of Technology Madras. He has contributed to interdisciplinary research with over 25 publications in journals and edited volumes by Elsevier, Springer, IEEE, Wiley, and CRC Press. His research interests include artificial intelligence, machine learning, 6G communications, healthcare technologies, and industrial automation.
}

\AuthorBio{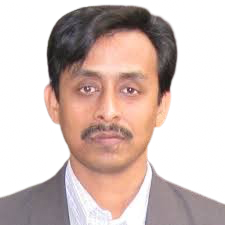}{Dr. Tapas Samanta}{
is a senior scientist and Head of the Computer and Informatics Group at the Variable Energy Cyclotron Centre (VECC), Department of Atomic Energy, India. With over two decades of experience, his work spans artificial intelligence, industrial automation, embedded systems, high-performance computing, and accelerator control systems. He also leads technology transfer initiatives and public scientific outreach at VECC.
}

\AuthorBio{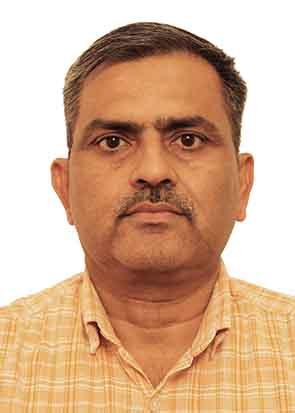}{Hemendra Kumar Pandey}{is a Scientific Officer in the Radioactive Ion Beam Facilities Group at the Variable Energy Cyclotron Centre (VECC), Department of Atomic Energy, Kolkata, India. He received his Ph.D. from the Indian Institute of Technology Kharagpur and his M.Tech. from the University of Allahabad. He joined Bhabha Atomic Research Centre in 1999 and has been associated with VECC since 2000, where he has contributed to RF and microwave systems for particle accelerators, including development activities for the Radioactive Ion Beam facility. He is also an Associate Professor at Homi Bhabha National Institute. His research interests include RF systems for particle accelerators, beam diagnostics, high-power RF amplifier development, mixed-signal RF integrated-circuit design, and radiation-hardened devices.in accelerator-based technologies.
}

\AuthorBio{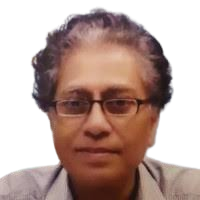}{Dr. Amitabha Das}{
is the Director and Head of the School of Nuclear Studies and Application at Jadavpur University, Kolkata. His research interests include nuclear instrumentation, embedded systems, reactor control systems, and FPGA-based real-time data acquisition. He has also contributed to AI-driven applications such as lipreading and sign language recognition and has supervised advanced research in nuclear reactor control methodologies.
}

\AuthorBio{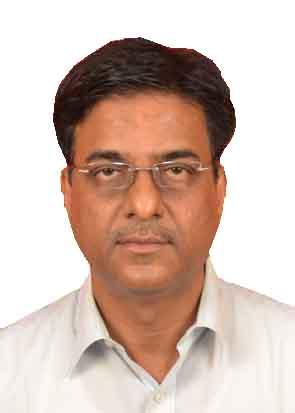}{Dr. Sarbajit Pal}{
is a retired senior scientist and former Head of the C\&I Group at the Variable Energy Cyclotron Centre (VECC), Department of Atomic Energy, Government of India. He holds a Ph.D. in Electronics Engineering and has made significant contributions to control and instrumentation systems for particle accelerators, including the K500 Superconducting Cyclotron. His expertise includes embedded systems, experimental physics, and EPICS-based control architectures.
}

\end{document}